\documentclass[
]{ceurart}

\usepackage{listings}

\usepackage{longtable}
\usepackage{array}
\usepackage{natbib}
\usepackage{textcomp}
\usepackage{float}

\begin{document}

\copyrightyear{2022}
\copyrightclause{Copyright for this paper by its authors.
  Use permitted under Creative Commons License Attribution 4.0
  International (CC BY 4.0).}

\conference{Woodstock'22: Symposium on the irreproducible science,
  June 07--11, 2022, Woodstock, NY}

\title{Semantic Analysis of SNOMED CT Concept Co-occurrences in Clinical Documentation using MIMIC-IV}

\author[1]{Ali Noori}[
email=a_noori@uncg.edu]
\address[1]{Informatics and Analytics,
  University of North Carolina Greensboro, Greensboro, NC}

\author[2]{Somya Mohanty}[%
email=mohanty.somya@gmail.com
]

\address[2]{United HealthGroup, United States}

\author[3]{Prashanti Manda}[%
email=pmanda@nebraska.edu
]
\cormark[1]
\address[3]{Department of Computer Science, University of Nebraska Omaha, Omaha, NE}
\begin{abstract}
Clinical notes contain rich clinical narratives but their unstructured format poses challenges for large-scale analysis. Standardized terminologies such as SNOMED CT improve interoperability, yet understanding how concepts relate through co-occurrence and semantic similarity remains underexplored. In this study, we leverage the MIMIC-IV database to investigate the relationship between SNOMED CT concept co-occurrence patterns and embedding-based semantic similarity. Using Normalized Pointwise Mutual Information (NPMI) and pretrained embeddings (e.g., ClinicalBERT, BioBERT), we examine whether frequently co-occurring concepts are also semantically close, whether embeddings can suggest missing concepts, and how these relationships evolve temporally and across specialties. Our analyses reveal that while co-occurrence and semantic similarity are weakly correlated, embeddings capture clinically meaningful associations not always reflected in documentation frequency. Embedding-based suggestions frequently matched concepts later documented, supporting their utility for augmenting clinical annotations. Clustering of concept embeddings yielded coherent clinical themes (symptoms, labs, diagnoses, cardiovascular conditions) that map to patient phenotypes and care patterns. Finally, co-occurrence patterns linked to outcomes such as mortality and readmission demonstrate the practical utility of this approach. Collectively, our findings highlight the complementary value of co-occurrence statistics and semantic embeddings in improving documentation completeness, uncovering latent clinical relationships, and informing decision support and phenotyping applications.
\end{abstract}


\maketitle

\section{Introduction}
Clinical notes are a rich repository of patient information, documenting observations, diagnoses, treatments, and outcomes across the continuum of care. They are critical for clinical decision-making, quality improvement, and research, but their unstructured format and scale make systematic analysis challenging \cite{wang2020clinical}. Efficient methods are needed to extract meaningful insights from this data. Standardized medical terminologies such as SNOMED CT \cite{donnelly2006snomed} improve interoperability and semantic clarity, providing a foundation for automated analysis. However, the complexity of clinical data persists: identifying patterns, understanding semantic relationships, and uncovering clinically relevant associations remain difficult. 

Advances in natural language processing (NLP) and embeddings (e.g., ClinicalBERT \cite{alsentzer2019clinicalbert}, BioBERT \cite{lee2020biobert}, SapBERT \cite{liu2021sapbert}) capture semantic nuances of biomedical text, enabling similarity assessments and inference. Yet their integration with standardized terminologies like SNOMED CT in large-scale clinical data is still limited.  

One promising direction is analyzing co-occurrence of SNOMED CT concepts in clinical notes and relating these patterns to semantic similarity. Co-occurrences may reflect diagnostic reasoning, symptom–condition associations, or atypical presentations. Frequent co-occurrence of semantically similar concepts can reinforce known pathways, while unexpected pairings may reveal undocumented phenotypes, inconsistencies, or missing concepts. Such analyses can enhance EHR completeness, support billing and population health, and uncover patient subgroups for phenotyping.  

This study leverages the MIMIC-IV database \cite{johnson2023mimiciv} to annotate clinical notes with SNOMED CT concepts and investigate semantic relationships in co-occurrence patterns. We address key questions: Do frequently co-occurring concepts also cluster semantically? Can semantic similarity suggest missing concepts? Do embedding-based relationships align with clinical knowledge and reveal novel subgroups?  

Our contributions are:
\begin{enumerate}
\item A large-scale analysis of SNOMED CT concept co-occurrences and embedding-based similarities.
\item Embedding-driven clustering to identify clinically meaningful semantic groups.
\item Exploration of temporal co-occurrence dynamics across hospital stays.
\item Comparative analysis across specialties and documentation types.
\item Identification of unusual but clinically relevant concept pairs.
\item Linking co-occurrence patterns to outcomes such as mortality, length-of-stay, and readmission.
\end{enumerate}

Collectively, these contributions aim to bridge the gap between standardized clinical terminologies, semantic embedding methodologies, and clinical informatics applications, thereby enhancing the utility and interpretability of clinical notes in real-world settings.

\section{Related Work}
Several studies have leveraged the MIMIC database in conjunction with SNOMED CT annotations to address various clinical informatics challenges. Johnson et al. \cite{johnson2016mimic} first introduced the MIMIC dataset, which has since become a foundational resource for research into intensive care and clinical documentation analytics. SNOMED CT annotations have been utilized extensively in analyzing clinical events, patient phenotypes, and outcomes within MIMIC datasets.

For instance, Sohn et al. \cite{sohn2018snomed} used SNOMED CT to study acute kidney injury within the MIMIC-III dataset. They applied association rule mining on SNOMED-annotated discharge summaries to uncover frequent concept combinations predictive of clinical deterioration. Similarly, Yu et al. \cite{yu2020mortality} utilized logistic regression and random forest models to predict patient mortality based on SNOMED-annotated clinical notes from MIMIC-III, demonstrating improved predictive performance compared to traditional structured data approaches.

Wang et al. \cite{wang2021clinicalbert} further expanded the scope of semantic analyses by integrating ClinicalBERT embeddings with SNOMED CT concepts derived from MIMIC-III notes. This combination enabled patient clustering based on semantic similarity, revealing distinct patient groups associated with varying clinical outcomes such as mortality and hospital readmission. Their findings underscore the potential of semantic embeddings combined with standardized terminologies to uncover meaningful clinical phenotypes.

Beam et al. \cite{beam2020clinical,beam2020multimodal} also leveraged SNOMED annotations in a multimodal embedding approach, integrating structured clinical data with textual notes to generate embeddings. These embeddings were subsequently analyzed using cosine similarity and hierarchical clustering, identifying clinically relevant semantic relationships and associations not readily apparent through conventional analyses.

Additional studies have explored temporal aspects of SNOMED CT annotations within MIMIC data. Miotto et al. \cite{miotto2016deep} examined temporal shifts in clinical concept usage during hospital stays, identifying distinct patterns that corresponded to evolving clinical presentations and interventions. These temporal patterns highlighted the dynamic nature of clinical documentation and underscored the importance of temporal considerations in semantic analysis.

Despite these advancements, the direct study of concept co-occurrence patterns within SNOMED annotations and their relationship to semantic similarity remains limited. While Wang et al.\cite{wang2021clinicalbert} and Beam et al. \cite{beam2020clinical} introduced semantic embeddings into their analysis pipelines, their focus was primarily on patient-level embedding aggregation and outcome prediction, rather than a detailed exploration of intra-note concept relationships or co-occurrence-based semantics. Moreover, few studies have explicitly linked high-frequency co-occurrence patterns to embedding-based similarity metrics, nor have they examined the potential of such an approach to identify missing concepts, documentation inconsistencies, or clinical subtypes.

To our knowledge, no prior work has systematically investigated the correlation between co-occurrence frequency (e.g., NPMI) and semantic embedding distance within SNOMED-annotated notes in MIMIC. Furthermore, the potential of this joint analysis to identify temporally dynamic patterns, specialty-specific associations, or predictive signals for patient outcomes remains an open research question. Our study addresses this gap by combining SNOMED-based co-occurrence analysis with embedding similarity metrics across multiple dimensions: temporal progression, clinical specialty, unusual concept pair detection, and outcome correlation. In doing so, we aim to offer a deeper understanding of how semantic and statistical properties of clinical concepts interact within real-world documentation and how this interaction can be harnessed for clinical insight and decision support.

\section{Methods}
\subsection{MIMIC-IV Clinical Notes}
The Medical Information Mart for Intensive Care IV (MIMIC-IV) database is a publicly available, de-identified dataset comprising detailed health-related information on patients admitted to intensive care units at the Beth Israel Deaconess Medical Center. One of the most valuable components of MIMIC-IV is its rich corpus of clinical notes, which are stored in the 	exttt{noteevents} table. These notes capture unstructured narratives authored by physicians, nurses, and other clinical staff, and are an essential source of qualitative information that complements structured data such as lab results, medications, and diagnoses.

The MIMIC-IV database includes over 400,000 clinical notes authored between 2008 and 2019, covering more than 40,000 distinct hospital admissions and approximately 70,000 patients. The clinical notes encompass a wide range of document types, including discharge summaries, progress notes, nursing notes, radiology reports, and emergency department documentation. Each note is timestamped and associated with metadata such as the type of caregiver and the category of the note (e.g., ``Discharge summary'', ``Physician'', ``Nursing''). This metadata enables researchers to stratify notes temporally or by clinical context.

Given their narrative nature, these notes often contain detailed explanations of patient conditions, rationale for clinical decisions, and implicit relationships among symptoms, diagnoses, and treatments. However, the unstructured format of clinical text poses significant challenges for computational analysis. To facilitate semantic processing, the notes are frequently preprocessed and mapped to standardized clinical ontologies such as SNOMED CT. This annotation process enables downstream tasks such as concept extraction, co-occurrence analysis, and semantic similarity computation.

In this study, we utilize the SNOMED CT-annotated version of these clinical notes to investigate the co-occurrence patterns of clinical concepts and their semantic proximity in embedding space. By examining how concepts appear together within and across notes, we aim to uncover the latent clinical logic and knowledge structure embedded in documentation practices.

\subsection{Data Preparation}
We extracted SNOMED CT concept co-occurrences from clinical notes in MIMIC-IV. Two SNOMED concepts were considered to co-occur if they appeared within the same clinical note. This notion of co-occurrence assumes that concepts mentioned together in a single note may be clinically related, either directly (e.g., symptom and diagnosis) or through shared clinical context (e.g., coexisting conditions).

\subsubsection{Strength of Co-occurrence}
To quantify the strength of co-occurrence, we computed Normalized Pointwise Mutual Information (NPMI), which measures how strongly two concepts are associated relative to their individual frequencies. NPMI is defined as:

\begin{equation}
\text{NPMI}(x,y) = \frac{\log \frac{p(x,y)}{p(x)p(y)}}{-\log p(x,y)}
\end{equation}

where \( p(x,y) \) is the joint probability of two concepts appearing in the same note, and \( p(x) \), \( p(y) \) are the marginal probabilities of each concept. NPMI values range from -1 to +1, where values closer to +1 indicate stronger-than-expected co-occurrence, values around 0 indicate independence, and values near -1 suggest mutual exclusivity.

\subsubsection{Semantic representation of SNOMED CT concepts}
To analyze semantic similarity, we generated vector representations (embeddings) of SNOMED CT concepts using pre-trained models such as ClinicalBERT and BioBERT. These embeddings encode contextual information about each concept and allow us to compute semantic distances—using metrics like cosine similarity or Euclidean distance—between co-occurring pairs. This dual perspective of co-occurrence strength and semantic similarity helps uncover whether concepts that frequently appear together are also semantically related, and can be used to detect latent clinical patterns or suggest missing documentation.


\subsubsection{Temporal Co-occurrence Analysis}
To investigate how concept co-occurrences evolve over the course of hospital stays, we performed a temporal segmentation of clinical notes in MIMIC-IV and analyzed differences in concept relationships between early and late stages of hospitalization. This approach was designed to capture dynamic shifts in clinical focus, such as initial symptoms and diagnoses versus downstream complications or discharge planning.

\subsubsection{Temporal Segmentation of Notes and identifying Co-occurring concept pairs}
To refine the segmentation of notes beyond fixed time thresholds, we calculated the total duration of each patient’s hospital stay by subtracting the admission time (\texttt{admittime}) from the discharge time (\texttt{dischtime}). Each clinical note was then assigned a label based on its timing relative to this duration. Specifically, we computed the delay between the note's chart time (\texttt{charttime}) and the admission time, and labeled a note as ``early'' if it was authored within the first 70\% of the hospital stay, and ``late'' otherwise. This approach provides a patient-specific, normalized temporal framework that accommodates variable lengths of stay and allows for consistent comparison of early and late-stage documentation patterns across patients. A bar chart of the distribution of early vs. late notes was generated to validate the segmentation logic.

Each clinical note was preprocessed to identify SNOMED CT concepts. The identified concepts from each note were then mapped to their respective time window (early or late) based on the note’s timestamp.

Within each time window, we extracted all unique unordered pairs of co-occurring SNOMED CT concepts per patient note. Two concepts were considered to co-occur if they appeared in the same note. We aggregated these across all patient notes to construct co-occurrence matrices for both early and late windows.

\subsubsection{NPMI Computation and Comparison}
We computed Normalized Pointwise Mutual Information (NPMI) separately for the early and late co-occurrence matrices to quantify the association strength between concept pairs. NPMI is defined as:
\begin{equation}
\text{NPMI}(x,y) = \frac{\log \frac{p(x,y)}{p(x)p(y)}}{-\log p(x,y)}
\end{equation}
where \( p(x,y) \) is the joint probability of concepts \( x \) and \( y \) appearing together in a note, and \( p(x) \), \( p(y) \) are their respective marginal probabilities.

\subsubsection{Identifying Temporal Shifts}
To identify meaningful temporal differences, we compared the NPMI scores between the early and late windows. We flagged:
\begin{itemize}
  \item Concepts that appear exclusively in early or late stages.
  \item Concept pairs with large absolute differences in NPMI values (\( \Delta\text{NPMI} \)).
\end{itemize}

\section{Results}

\begin{figure}[]
    \centering
    \includegraphics[width=0.75\textwidth]{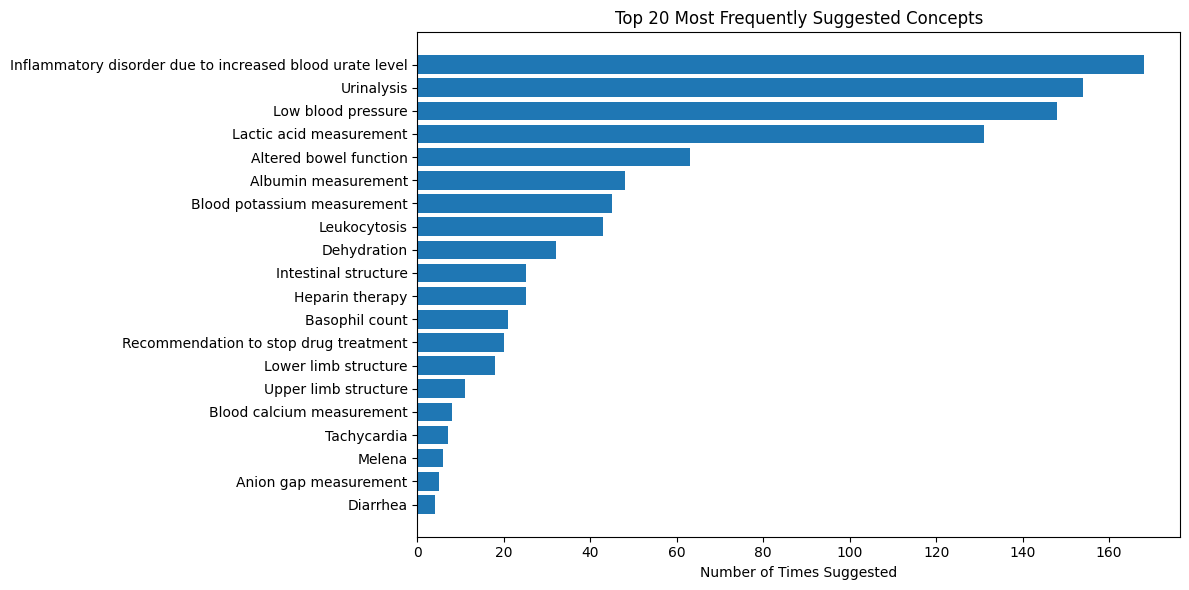}
    \caption{Top 20 most frequently suggested SNOMED CT concepts across all clinical notes. The x-axis indicates the number of times each concept was suggested for annotation, and the y-axis lists the corresponding clinical concept descriptions.}
    \label{fig:11}
\end{figure}

Figure~\ref{fig:11} shows the 20 most frequently suggested SNOMED CT concepts across all clinical notes. Laboratory and diagnostic observations dominate, led by \textit{Inflammatory disorder due to increased blood urate level} (over 160 mentions), reflecting the frequent discussion of conditions such as gout. Other common suggestions include \textit{Urinalysis}, \textit{Low blood pressure}, and \textit{Lactic acid measurement}, indicating the model’s focus on lab tests and vital signs central to intensive care. Similarly, concepts like \textit{Albumin measurement}, \textit{Blood potassium measurement}, and \textit{Dehydration} emphasize its sensitivity to acute physiological markers. In contrast, anatomical and procedural terms (e.g., \textit{Lower limb structure}, \textit{Intestinal structure}) appear less often, suggesting more context-dependent relevance. Therapeutic concepts such as \textit{Heparin therapy} and \textit{Recommendation to stop drug treatment} demonstrate attention to treatment decisions, while less frequent but diagnostically important terms like \textit{Melena}, \textit{Tachycardia}, and \textit{Diarrhea} highlight the model’s ability to capture critical but rarer signals. Overall, the figure illustrates how the model prioritizes common measurements and interventions while still surfacing clinically meaningful low-frequency concepts.

\begin{figure}[]
    \centering
    \includegraphics[width=0.75\textwidth]{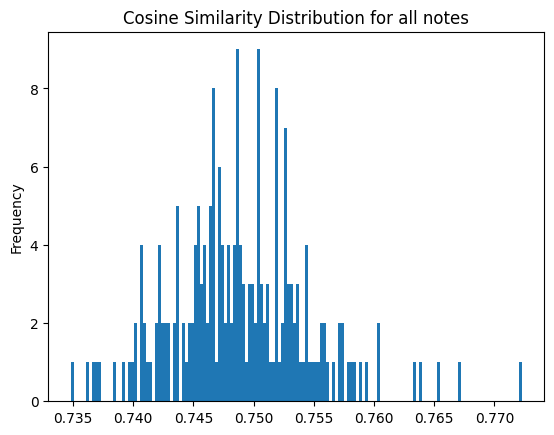}
    \caption{Histogram showing the distribution of cosine similarity scores between SNOMED CT concept pairs across all clinical notes in MIMIC-IV. Cosine similarity was computed using concept embeddings derived from a pretrained model (e.g., BioBERT or ClinicalBERT), capturing the semantic proximity between concept pairs.}
    \label{fig:1}
\end{figure}

Next, we examine similarity between co-occurring SNOMED CT concepts in clinical notes. Figure~\ref{fig:1} shows that cosine similarity scores cluster around 0.75, indicating most pairs are moderately to highly related and often form coherent clinical narratives (e.g., \textit{fever}–\textit{pneumonia}, \textit{intubation}–\textit{mechanical ventilation}). The right-skewed distribution suggests embeddings capture real-world clinical usage, though a small set of low-similarity pairs (<0.74) reflect disparate events such as unrelated symptoms or comorbidities. These outliers may reveal unusual associations, documentation inconsistencies, or complex presentations. Overall, the tight clustering underscores the embedding model’s alignment with clinical co-occurrence patterns and supports its use in detecting missing concepts and signals of clinical complexity.

\begin{figure}[]
    \centering
    \includegraphics[width=0.75\textwidth]{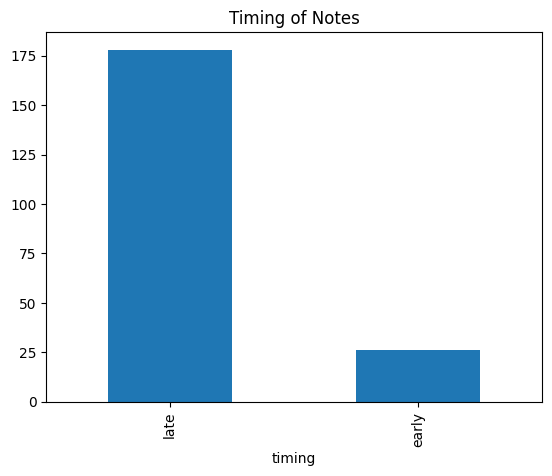}
    \caption{Bar chart showing the distribution of clinical notes labeled as ``early'' or ``late'' based on their timing relative to the total length of a hospital stay. Notes were labeled ``early'' if authored in the first 70\% of the stay, and ``late'' otherwise.}
    \label{fig:5}
\end{figure}

Clinical notes were classified as \textit{early} or \textit{late} based on timing within a patient’s hospital stay: the first 70\% of the stay was labeled \textit{early}, and the remainder \textit{late}. As shown in Figure~\ref{fig:5}, documentation is heavily skewed toward the late stage, with nearly seven times more notes than in the early group. This reflects real-world practice, where early notes are limited to admission and triage, while late notes contain frequent updates, complication tracking, and discharge planning. The imbalance has implications for temporal analysis, necessitating normalization across stages and underscoring the difficulty of extracting robust early-stage signals compared to the richer information available later.

\begin{figure}[]
    \centering
    \includegraphics[width=0.75\textwidth]{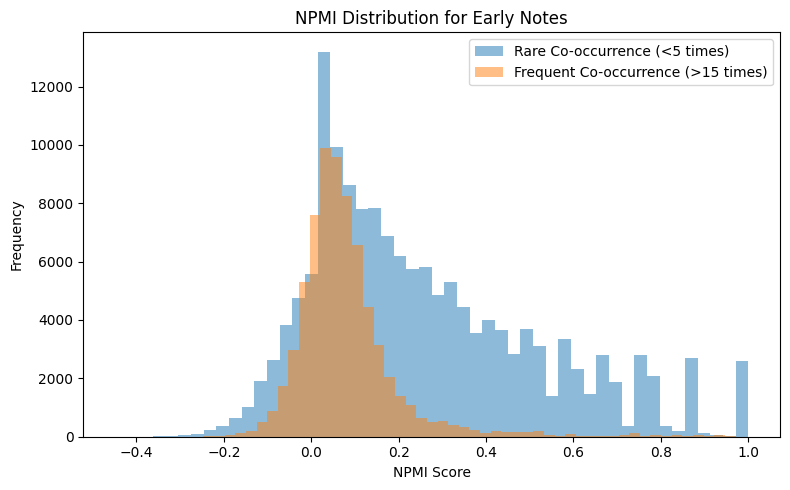}
    \caption{Histogram of Normalized Pointwise Mutual Information (NPMI) scores for SNOMED CT concept pairs in early-stage clinical notes. The distribution is stratified by co-occurrence frequency: rare co-occurrence (fewer than 5 notes) and frequent co-occurrence (more than 15 notes).}
    \label{fig:2}
\end{figure}

We then computed NPMI scores for co-occurring SNOMED CT concepts, separating pairs from early vs. late notes. Figure~\ref{fig:2} shows distributions for rare pairs (fewer than 5 co-occurrences, blue) and frequent pairs (more than 15, orange). Rare pairs cluster near zero, suggesting chance associations or documentation noise, while frequent pairs are skewed right, reflecting strong clinical relationships such as symptom–diagnosis links. Notably, some rare pairs also show high NPMI values, pointing to underdocumented but clinically meaningful associations or niche subgroup patterns. A few extreme outliers (NPMI > 0.8) represent canonical links (e.g., \textit{shortness of breath}–\textit{oxygen therapy}) or possible annotation artifacts. Overall, NPMI effectively distinguishes meaningful associations from spurious ones, demonstrating that frequency alone does not determine clinical relevance.

\begin{figure}[]
    \centering
    \includegraphics[width=0.75\textwidth]{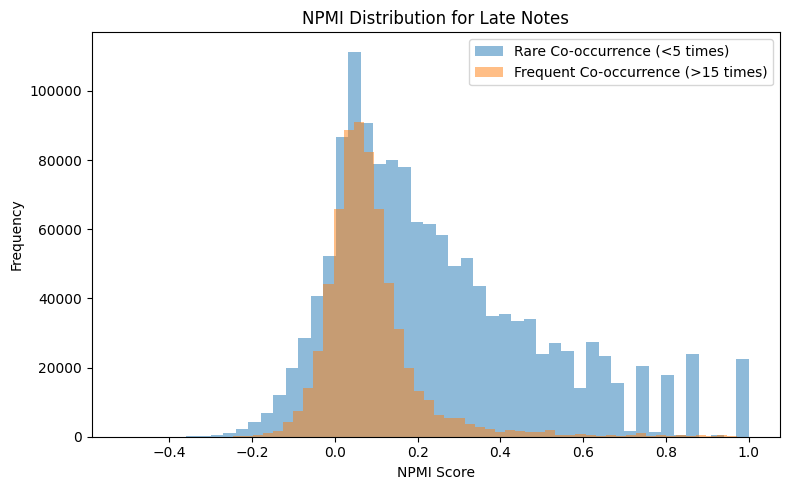}
    \caption{Histogram of Normalized Pointwise Mutual Information (NPMI) scores for SNOMED CT concept pairs in late-stage clinical notes. The distribution is separated by co-occurrence frequency, showing rare co-occurrences (fewer than 5 notes, in blue) and frequent co-occurrences (more than 15 notes, in orange).}
    \label{fig:4}
\end{figure}

We analyzed NPMI scores for concept pairs in the final 30\% of hospital stays (late notes), comparing rare (blue) and frequent (orange) co-occurrences (Figure~\ref{fig:4}). Both groups peak near zero, with a sharper spike for rare pairs, indicating weak or chance associations. Frequent pairs, however, show a stronger rightward skew than in early notes, with more exceeding 0.2, reflecting stable clinical relationships common in discharge planning or chronic care. Rare pairs also display a right tail, suggesting underdocumented but meaningful associations or rare subgroup patterns. A small fraction of pairs have negative scores, pointing to mutually exclusive events or distinct care phases. Overall, late notes contain more high-frequency, high-association pairs, underscoring how concept relationships strengthen over the course of care and highlighting the utility of NPMI for temporal analysis of clinical documentation.

\begin{figure}[]
    \centering
    \includegraphics[width=0.750\textwidth]{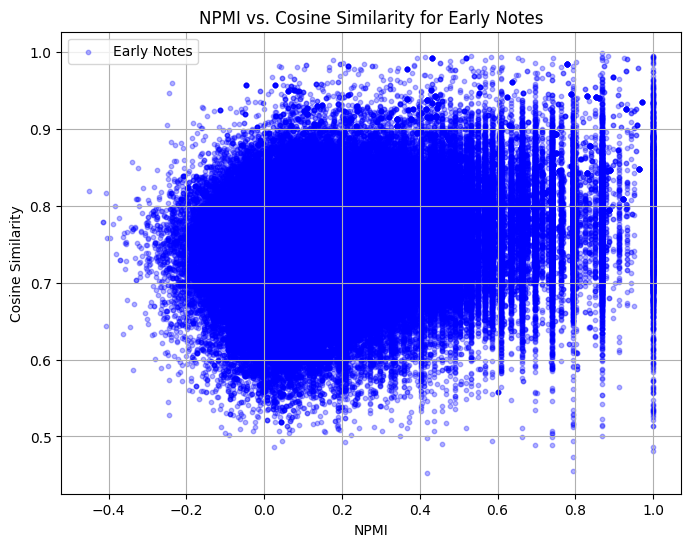}
    \caption{Scatter plot illustrating the relationship between Normalized Pointwise Mutual Information (NPMI) and cosine similarity for SNOMED CT concept pairs extracted from early clinical notes in MIMIC-IV. Each point represents a unique concept pair.}
    \label{fig:6early}
\end{figure}

\begin{figure}[]
    \centering
    \includegraphics[width=0.750\textwidth]{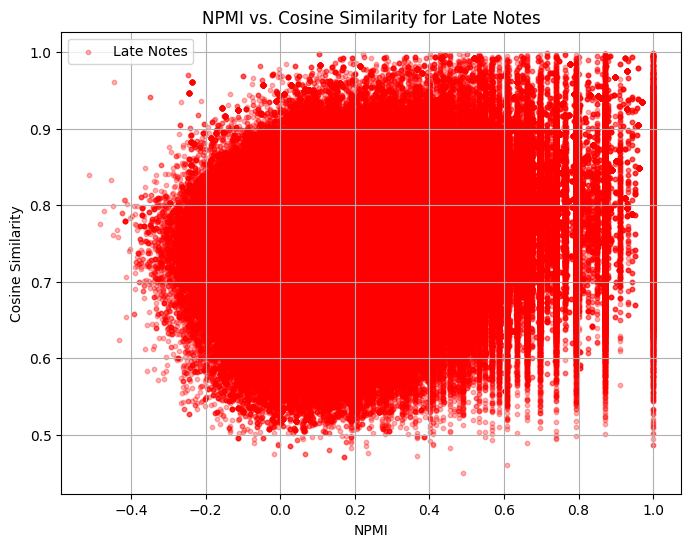}
    \caption{Scatter plot illustrating the relationship between Normalized Pointwise Mutual Information (NPMI) and cosine similarity for SNOMED CT concept pairs extracted from late clinical notes in MIMIC-IV. Each point represents a unique concept pair.}
    \label{fig:6late}
\end{figure}

Figures \ref{fig:6early} and \ref{fig:6late} illustrate the relationship between normalized pointwise mutual information (NPMI) and cosine similarity of concepts in early versus late notes. In both cases, cosine similarity values are generally high (centered between 0.7 and 0.9), while NPMI values show a wider spread ranging from negative to positive associations. The early notes (blue) display a denser clustering in the mid-range of NPMI (around 0 to 0.4) with a slightly more diffuse spread toward negative NPMI values, suggesting that concept pairs in early notes often share semantic similarity even when their co-occurrence is weak or negative. By contrast, the late notes (red) show a more compact and symmetric distribution with fewer extreme outliers, particularly at low cosine similarity values, indicating stronger alignment between co-occurrence and semantic similarity. Overall, while both note types show that high cosine similarity can occur across a wide range of NPMI values, early notes exhibit more variability, whereas late notes appear more consistent and concentrated. While the two metrics are positively associated, their differences also carry rich information that can help uncover latent clinical structures or documentation idiosyncrasies. This underscores the value of leveraging both co-occurrence statistics and embedding-based semantic measures in tandem when analyzing clinical concept relationships.

\begin{figure}[]
    \centering
    \includegraphics[width=0.75\textwidth]{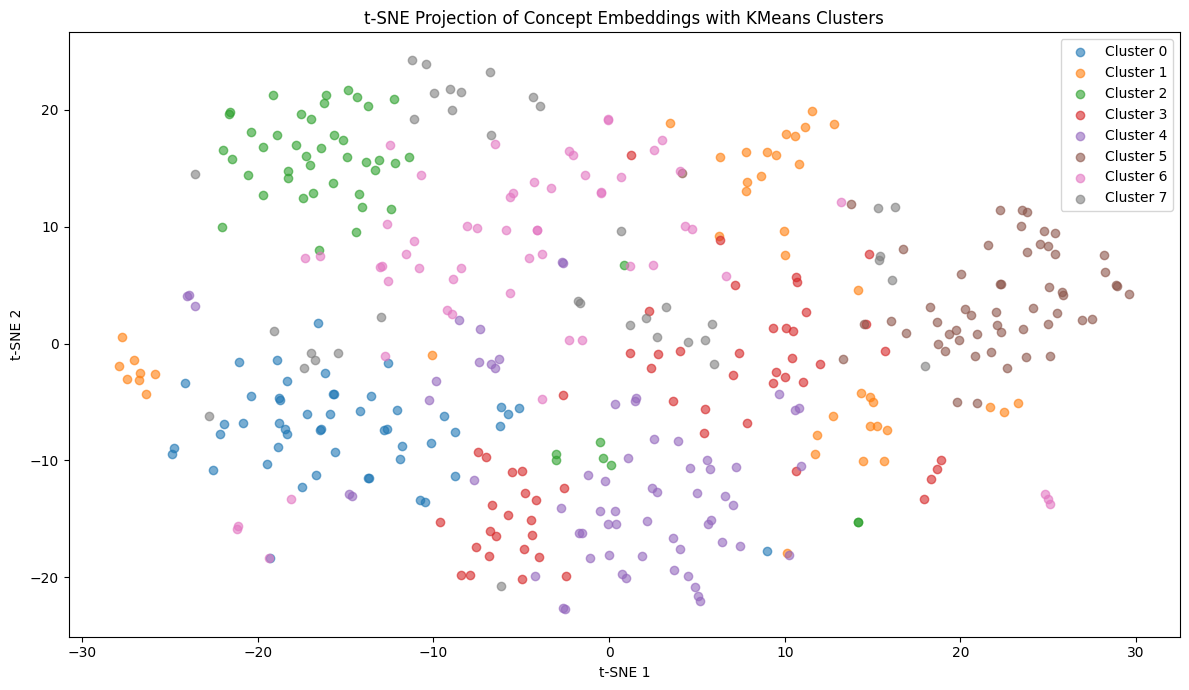}
    \caption{t-SNE projection of SNOMED CT concept embeddings annotated from MIMIC-IV clinical notes, color-coded by KMeans clustering (k=8). Each point represents an individual SNOMED concept projected into a 2D space from high-dimensional embedding vectors. Clusters reflect semantic groupings inferred from embedding similarity.}
    \label{fig:8}
\end{figure}

The concept clusters in Table~\ref{clustertable} reveal clear semantic organization consistent with clinical knowledge, underscoring the effectiveness of embedding-based grouping.

\textbf{Cluster 0: Anatomical Structures} – Terms such as \textit{Left atrial structure}, \textit{Right lung structure}, and \textit{Lower limb structure} reflect SNOMED CT’s ontological hierarchy and frequently co-occur in exam and imaging notes.

\textbf{Cluster 1: Examinations and Procedures} – Includes \textit{Examination of respiratory system}, \textit{Computed tomography}, and \textit{Administration of analgesic}, capturing common diagnostic and procedural language.

\textbf{Cluster 2: Symptoms and Complaints} – Concepts like \textit{Backache}, \textit{Dizziness}, and \textit{Vomiting} anchor diagnostic reasoning and often appear in triage and early documentation.

\textbf{Cluster 3: Diseases and Interventions} – A mix of diagnoses (\textit{Asthma}, \textit{Hematochezia}), treatments (\textit{Recommendation to stop drug treatment}), and patient factors (\textit{Physical functional dependency}), reflecting complex inpatient management.

\textbf{Cluster 4: Cardiovascular Conditions} – Includes \textit{Myocardial infarction}, \textit{Hypercholesterolemia}, and \textit{Coronary artery bypass grafting}, a highly coherent domain-specific grouping.

\textbf{Cluster 5: Lab Tests and Monitoring} – Terms such as \textit{Platelet count}, \textit{Sodium measurement}, and \textit{Blood pressure monitoring}, reflecting clinical monitoring workflows.

\textbf{Cluster 6: Status and Observations} – General assessments like \textit{Patient’s condition stable}, \textit{Warm skin}, and \textit{Normal vital signs}, found across specialties.

\textbf{Cluster 7: Miscellaneous Findings} – A heterogeneous mix (\textit{Fever}, \textit{Wound}, \textit{Kidney structure}, \textit{Problem}), representing ambiguous or cross-cutting concepts.

Overall, the clusters align with established clinical categories (e.g., cardiovascular, labs, symptoms) while also highlighting domains with greater semantic overlap. This organization supports applications in phenotyping, ontology refinement, and decision support.

\begin{longtable}{|c|p{3.5cm}|p{9.5cm}|}
\hline
\label{clustertable}
\textbf{Cluster} & \textbf{Cluster Name} & \textbf{Sample Concepts} \\
\hline
0 & Anatomical Structures and Normal Findings & Left atrial structure, Right lung structure, Pleural effusion, Structure of base of lung, Normal head, Fully mobile, Lower limb structure, Normal tissue perfusion, Right atrial structure, Blood vessel structure \\
\hline
1 & Procedures, Examinations, and Care Activities & Examination of respiratory system, Lying in bed, Plain X-ray of chest, Propensity to adverse reactions to drug, Application of dressing, Eruption of skin, Examination of genitourinary system, Administration of analgesic, Computed tomography, Monitoring for signs and symptoms of infection \\
\hline
2 & Symptoms and Patient-Reported Issues & Backache, Dizziness, Numbness, Decrease in appetite, Cough, Abdominal pain, Recent weight loss, Severe pain, Tenderness, Vomiting \\
\hline
3 & Chronic Conditions and Treatment Regimens & Asthma, Recommendation to stop drug treatment, Physical functional dependency, Hematochezia, Surgical incision wound, Inflammatory disorder, Nasal discharge, Diuretic therapy, Therapeutic regimen, Aspiration pneumonia \\
\hline
4 & Cardiovascular and Critical Events & Orthostatic body position, Transthoracic echocardiography, Mild mitral valve regurgitation, Cerebrovascular accident, Hypercholesterolemia, Myocardial infarction, Stenosis, Normal jugular venous pressure, Coronary artery bypass grafting, Pulmonary embolism \\
\hline
5 & Measurements and Interventions & Inflammatory disorder due to increased blood urate level, Bilirubin measurement (urine), Blood pressure monitoring, Platelet count, Lactic acid measurement, Sodium measurement, Oxygen administration by nasal cannula, Creatinine measurement, Urinalysis (acetone or ketone bodies), pH measurement \\
\hline
6 & General Observations and Vital Status & Patient's condition stable, Warm skin, Light touch sensation present, Normal vital signs, Cardiomegaly, Cyanosis, Normal abdominal contour, Normal movement of neck, Ecchymosis, Patient's condition worsened \\
\hline
7 & Generic or Ambiguous Findings & Kidney structure, Wound, Mood finding, Weight finding, Bleeding, Color finding, Heart structure, Problem, Fever, Dead \\
\hline
\end{longtable}

Our first research question asked whether frequently co-occurring SNOMED CT concepts also show high embedding similarity. Results indicate only a weak relationship: while some symptom–treatment pairs are both frequent and semantically close, many highly similar pairs rarely co-occur, showing that embeddings capture latent clinical associations beyond documentation frequency.

The second question explored whether semantic similarity can suggest missing concepts. By averaging embeddings of documented terms and retrieving nearby concepts, we generated contextually appropriate suggestions, many of which later appeared in follow-up notes. This demonstrates potential for improving documentation completeness and supporting decision support or coding automation. The third question asked if embedding-based similarity reflects clinically meaningful relationships. We found that many high-similarity pairs align with known diagnostic, procedural, or physiological links—even when absent from the SNOMED CT hierarchy—indicating that embeddings capture contextual relationships consistent with real-world practice. Finally, clustering embeddings revealed coherent clinical themes (e.g., symptoms, lab tests, anatomical structures, diagnoses) that map to patient-level phenotypes and care patterns. This suggests embeddings can enrich patient stratification, ontology refinement, and nuanced clinical analysis. Overall, semantic similarity and co-occurrence provide complementary perspectives on documentation and patient trajectories. Embedding methods effectively identify clinically relevant relationships, support missing concept suggestion, and uncover potential anomalies or novel associations. Future work should include prospective validation and extend to broader embedding approaches and clinical contexts.

\begin{acknowledgments}
This work is funded by a CAREER award (\#2522386) to Manda from the Division of Biological Infrastructure at the National Science Foundation, USA.
\end{acknowledgments}

\bibliography{references}

\end{document}